\documentclass[twoside]{article}

\usepackage[accepted]{aistats2023}
\usepackage[round]{natbib}

\usepackage{xcolor,xr-hyper,xr}
\usepackage{hyperref,float}   
\usepackage{subcaption}
\usepackage{algorithm,enumerate}
\usepackage{algpseudocode,bm}
\usepackage{amsfonts,  amsmath}       
\usepackage{comment}
\usepackage{graphicx}

\def\Ibf{I}
\def\wbf{w}
\def\phibf{\phi}

\newtheorem{assump}{Assumption}

\newtheorem{theorem}{Theorem}

\newtheorem{corollary}{Corollary}
\newtheorem{proposition}{Proposition}

\begin{document}

%

%

\twocolumn[

\aistatstitle{Optimizing Pessimism in Dynamic Treatment Regimes: A Bayesian Learning Approach}

\aistatsauthor{ Yunzhe Zhou \And Zhengling Qi  \And  Chengchun Shi \And Lexin Li}

\aistatsaddress{ UC Berkeley \And  George Washington University \And LSE \And UC Berkeley} ]

\begin{abstract}
In this article, we propose a novel pessimism-based Bayesian learning method for optimal dynamic treatment regimes in the offline setting. When the coverage condition does not hold, which is common for offline data, the existing solutions would produce sub-optimal policies. The pessimism principle addresses this issue by discouraging recommendation of actions that are less explored conditioning on the state. However, nearly all pessimism-based methods rely on a key hyper-parameter that quantifies the degree of pessimism, and the performance of the methods can be highly sensitive to the choice of this parameter. We propose to integrate the pessimism principle with Thompson sampling and Bayesian machine learning for optimizing the degree of pessimism. We derive a credible set whose boundary uniformly lower bounds the optimal Q-function, and thus we do not require additional tuning of the degree of pessimism. We develop a general Bayesian learning method that works with a range of models, from Bayesian linear basis model to Bayesian neural network model. We develop the computational algorithm based on variational inference, which is highly efficient and scalable. We establish the theoretical guarantees of the proposed method, and show empirically that it outperforms the existing state-of-the-art solutions through both simulations and a real data example.  
\end{abstract}

\section{INTRODUCTION}

Due to heterogeneity in patients' responses to the treatment, one-size-fits-all strategy may no longer be optimal \citep{jiang2017estimation}. Precision medicine aims to identify the most effective treatment strategy based on individual patient information.  For example, for many complex diseases, such as cancer, mental disorders and diabetes, patients are usually treated at multiple stages over time based on their evolving treatment and clinical covariates  \citep{sinyor2010sequenced, maahs2012outpatient}. Dynamic treatment regimes (DTRs) provide a useful framework of leveraging data to learn the optimal treatment strategy by incorporating heterogeneity across patients and time \citep{murphy2003optimal}. Formally, a DTR is a sequence of decision rules, where each rule takes the patient's past information as input, and outputs the treatment assignment. An optimal DTR is the one that maximizes patient's expected clinical outcomes. DTRs generally follow an \textit{online} learning paradigm, where the process involves repeatedly collecting patient's response to the assigned treatment. In medical studies, however, it is often impractical to constantly collect such interactive information. This prompts us to study the problem of learning optimal DTRs in an \textit{offline} setting, where the data have already been pre-collected. In this article, we propose a novel Bayesian learning approach using a pessimistic-type Thompson sampling for finding DTRs.

\subsection{Related Work}

\textbf{Statistical methods for DTRs}. There is a vast literature on statistical methods for finding optimal DTRs, which, broadly speaking, includes Q-learning, A-learning and value search methods. See \citet{tsiatis2019dynamic,kosorok2019precision} for an overview. See also \citet{robins2004optimal,qian2011performance,zhang2013robust,chakraborty2014dynamic,zhao2015new,chen2016personalized,shi2018high,shi2018maximin,qi2020multi,chen2020representation,zhang2020designing,cai2021jump,qiu2021optimal,zhou2021parsimonious,qi2022proximal}, and the references therein. However, most existing methods rely on a positivity assumption in the offline data, which essentially requires the probability of each treatment assignment at each stage is uniformly bounded away from zero. In the observational data, such an assumption could easily fail, as certain treatments are prohibited in some scenarios. Therefore, applying these methods may produce sub-optimal DTRs.

\textbf{Offline reinforcement learning (RL)}. Built on Markov decision process (MDP), Offline RL learns an optimal policy from historical data without any online interaction \citep{prudencio2022survey}. It is thus highly relevant for precision medicine type applications. However, many RL algorithms rely on a crucial coverage assumption, which requires the offline data distribution to provide a good coverage over the state-action distribution induced by all candidate policies. This assumption may be too restrictive and may not hold in observational studies. To address this challenge, the pessimism principle has been adopted that discourages recommending actions that are less explored conditioning on the state. The solutions in this family can be roughly classified into two categories, including model-based algorithms \citep[see e.g.,][]{kidambi2020morel,yu2020mopo,uehara2021pessimistic,yin2021near}, and model-free  algorithms \citep[see e.g.,][]{fujimoto2019off,kumar2019stabilizing,wu2019behavior,buckman2020importance,kumar2020conservative,rezaeifar2021offline,jin2021pessimism,xie2021bellman,zanette2021provable,bai2022pessimistic,fu2022offline}. The main idea of the model-based solutions is to penalize the reward or transition function whose state-action pair is rarely seen in the offline data, whereas the main idea of the model-free ones is to learn a conservative Q-function that lower bounds the oracle Q-function. Nevertheless, most of these solutions either require a well-specified parametric model, or rely on a key  hyperparameter to quantify the degree of pessimism. It is noteworthy that the performance of those solutions can be highly sensitive to the choice of the hyperparameter; see Section \ref{subsec:pess} for more illustration.  In addition, many algorithms are developed in the context of long or infinite-horizon Markov decision process. Their generalizations to medical applications with non-Markovian and finite-horizon systems remain unknown. Finally, we note that there is concurrent work by \cite{jeunen2021pessimistic} that adopts a Bayesian framework for offline contextual bandit. However, their method requires linear function approximations, and cannot handle complex nonlinear systems, nor more general sequential decision making.

\textbf{Thompson sampling}. Thompson sampling (TS) is a popular Bayesian approach proposed by \citet{thompson1933likelihood} that randomly draws each arm according to its  probability of being optimal, so to balance the exploration-exploitation trade-off in the online contextual bandit problems. It has demonstrated a competitive performance in empirical applications. For instance, \cite{chapelle2011empirical} showed that TS outperforms the upper confidence bound (UCB) algorithm in both synthetic and real data applications of advertisement and news article recommendation. The success of TS can be attributed to the Bayesian framework it adopts. In particular, the prior distribution serves as a regularizer to prevent overfitting, which  implicitly discourages exploitation. In addition, actions are selected randomly at each time step according to the posterior distribution, which explicitly encourages exploration and is useful in settings with delayed feedback \citep{chapelle2011empirical}.

\textbf{Bayesian machine learning}. Bayesian machine learning (BML) is a paradigm for   constructing machine learning models based on the Bayes theorem, and has been successfully deployed in a wide range of applications \citep[see, e.g.,][for a review]{seeger2006bayesian}. Popular BML methods include Bayesian linear basis model \citep{smith1973general}, variational autoencoder \citep{kingma2013auto}, Bayesian random forests \citep{quadrianto2014very}, Bayesian neural network \citep{blundell2015weight}, among many others. An appealing feature of BML is that, through posterior sampling, the uncertainty quantification is straightforward. In contrast, the frequentist methods for uncertainty quantification that are based on asymptotic theories can be highly challenging with complex machine learning models, whereas those based on bootstrap can be computationally intensive with large datasets.

\subsection{Our Proposal and Contributions}

In this article, we propose a novel pessimism-based Bayesian learning approach for offline optimal dynamic treatment regimes. We integrate the pessimism principle and Thompson sampling with the Bayesian machine learning framework. In particular, we derive an explicit and uniform uncertainty quantification of the Q-function estimator given the data, which in turn offers an alternative way of constructing confidence interval without having to specify a parametric model or tune the degree of pessimism, as required by nearly all existing pessimism-based offline RL and DTR algorithms. Compared to the RL and DTR algorithms without pessimism, our method yields a better decision rule when the coverage condition is seriously violated, and a comparable result when the coverage approximately holds. Compared to the RL and DTR algorithms adopting pessimism, our method achieves a more consistent and competitive performance. Theoretically, we show that the regret of the proposed method depends only on the estimation error of the \textit{optimal} action's Q-estimator, and we provide the explicit form of its upper bound in a special case of parametric model. The resulting bound is much narrower than the regret of the standard Q-learning algorithm that depends on the uniform estimation error of the Q-estimator at \textit{each} action. Methodologically, our approach is fairly general, and works with a range of different BML models, from simple Bayesian linear basis model to more complex Bayesian neural network model. Scientifically, our proposal offers a viable solution to a critical problem in precision medicine that can assist patients to achieve the best individualized treatment strategy. Finally, computationally, our algorithm is efficient and scalable to large datasets, as it adopts a variational inference approach to approximate the posterior distribution, and does not require computationally intensive posterior sampling method such as Markov chain Monte Carlo \citep{geman1984stochastic}.

Our method shares a similar spirit as TS, in that we also adopt a Bayesian framework for uncertainty quantification and exploration-exploitation trade-off. We also remark that, although the concepts of pessimism, TS and BML are not completely new, how to integrate them properly is highly nontrivial, and is the main contribution of this article. First of all, in the online setting, TS randomizes over actions to address the exploration-exploitation dilemma. However, randomization contradicts the pessimistic principle in the offline setting. To address this issue, we borrow the idea from the Bayesian UCB method \citep{kaufmann2012bayesian} for online bandits and generalize it to offline sequential decision making. Second, although posterior sampling allows one to conveniently quantify the \textit{pointwise} uncertainty of the estimated Q-function at a given individual state-action pair, it remains challenging 
to lower bound the Q-function \textit{uniformly} for any state-action pair. Developing a uniform credible set is crucial for implementing the pessimism principle. Our proposal provides an effective solution with a uniform uncertainty quantification.

\section{PRELIMINARIES}
\label{sec:moti}

\subsection{Bayesian Machine Learning} 
\label{subsec:bml}

Let $p(o|\wbf)$ denote a machine learning model indexed by $\wbf$ that parameterizes the probability mass or density function of some random variable $O$, and let $\mathcal{D}_n = \{o_i\}_{i=1}^n$ denote a set of i.i.d.\ random samples. BML treats $\wbf$ as a random quantity, and learns the entire posterior distribution $p(\wbf|\mathcal{D}_n)$ of $\wbf$ given the data $\mathcal{D}_n$  based on the Bayes rule, by combining the likelihood function $p(\mathcal{D}_n|\wbf)$ and a prior distribution $p(\wbf)$ that reflects prior knowledge about $\wbf$. Once the posterior distribution of $\wbf$ is learned, a commonly used point estimator for $\wbf$ is the posterior mean denoted by $\widehat \wbf = \mathbb{E}(\wbf \mid \mathcal{D}_n)$. One can then make the prediction by using $\widehat \wbf$ and the likelihood function. Alternatively, one can also make the prediction by using the posterior mean of the model output. We next consider two specific examples.

\textbf{Bayesian Linear Basis Model (BLBM)}. 
BLBM is an extension of the classical Bayesian linear model \citep{lindley1972bayes}, and models the distribution of a response $Y$ given $X=x$ as $y_i = \wbf^T \phibf(x_i) =  \sum_{j=1}^K w_j \phi_j(x_i) + \epsilon_i$, where $\phibf(x) = \{\phi_1(x),\cdots,\phi_K(x)\}^{\top}$ is a set of $K$ basis functions, $\wbf = (w_1,\ldots,w_n)^T$ is the weight vector, and the error $\epsilon_i$ follows a Gaussian distribution. Since the posterior distribution can be explicitly derived, 
BLBM is easy to implement in practice. However, it might suffer from potential model misspecification in high-dimensional complex problems.

\textbf{Bayesian Neural Network (BNN)}.
BNN learns the posterior distribution of the weight parameter $\wbf$ in a neural network. However, exact Bayesian inference is generally intractable due to the extremely complex model structure. \cite{blundell2015weight} proposed to approximate the exact posterior distribution $p(\wbf|\mathcal{D}_n)$ by a variational distribution $q(\wbf|\theta)$ whose functional form is pre-specified, and then estimate $\theta$ by minimizing the Kullback-Leibler (KL) divergence, $\text{KL}[q(\wbf | \theta)||p(\wbf|\mathcal{D}_n)]$. In practice, $q(\wbf|\theta)$ can be set to a multivariate Gaussian distribution, and the parameters are updated based on Monte Carlo gradients. \cite{blundell2015weight} developed an efficient computational algorithm, and showed BNN achieves a superior performance in numerous tasks.

\subsection{The Pessimism Principle} 
\label{subsec:pess}

In the offline setting, when the coverage condition is not met, the classical DTR and RL methods may yield sub-optimal policies. This is because some states and actions are less covered in the data, whose corresponding Q-values are difficult to learn, resulting in large variances and ultimately sub-optimal decisions. To address this issue, most existing offline RL methods adopt the pessimistic strategy, and derive the policies to avoid uncertain regions that are less covered in the data. Particularly, model-free offline RL methods learn a conservative  Q-estimator that lower bounds the Q-function during the search of the optimal policy. We next briefly review a state-of-the-art solution of this type, the pessimistic value iteration method (PEVI) of \citet{jin2021pessimism} based on linear models.

Consider a contextual bandit setting, where the offline data $\mathcal{D}_n$ consists of $n$ i.i.d.\ realizations $\{s_i,a_i,r_i\}_{i=1}^n$ of the state, action and reward tuple $\{S, A, R\}$, where $s_i$ collects the baseline covariates of the $i$th instance, $a_i$ is the action received, and $r_i$ is the corresponding reward. We assume $R$ is uniformly bounded and a larger value of $R$ indicates a better outcome. Denote the space of the covariates and actions by $\mathcal{S}$ and $\mathcal{A}$, respectively. In addition to estimating the conditional mean of the reward given the state-action pair, i.e., $Q(S,A)=\mathbb{E}(R|S,A)$, \citet{jin2021pessimism} proposed to also learn a $\xi$-uncertainty quantifier $\Gamma$, such that the event 
\begin{align}\label{eqn:condset}
\Omega = \Big\{|\widehat{Q}(s,a)-Q(s,a)| \leq \Gamma(s,a) \; \text{for all} \; (s,a) 
\Big\}    
\end{align}
holds with probability at least $1-\xi$ for any $\xi>0$, where $\widehat{Q}$ is an estimator of $Q$. Instead of computing the greedy policy with respect to $\widehat{Q}$ as in the standard methods, 
they proposed to choose the greedy policy that maximizes the lower bound $\widehat{Q}-\Gamma$, and showed that the regret of the resulting policy is upper bounded by $\mathbb{E}[\Gamma(S, \pi^\ast(S))]$, where $\pi^\ast$ is the true optimal policy. Note that this bound is much narrower than $\mathbb{E}[\max_a\Gamma(S, a)]$, i.e., the regret bound without taking pessimism into account. They further showed that the resulting policy is minimax optimal in linear finite horizon MDPs without the coverage assumption.

\begin{figure*}[t!]
\centering
\includegraphics[width=0.95\textwidth]{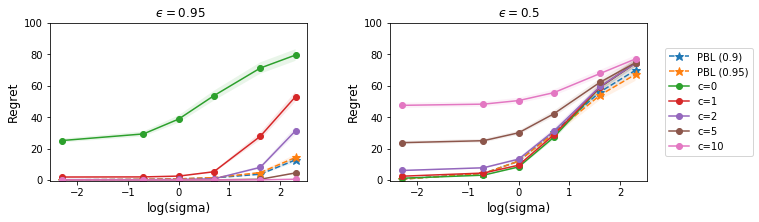}
\vspace{0.8cm}
\caption{A toy example comparing the PEVI method of \cite{jin2021pessimism} under different values of $c$ and our proposed method PBL.}
\label{fig:toy}
\end{figure*}

Despite its nice theoretical properties, it is challenging to implement PEVI in practice due to the construction of a proper $\Gamma$ that meets the requirement in \eqref{eqn:condset}. \cite{jin2021pessimism} only developed a construction of $\Gamma$ under a linear MDP model, and it cannot be easily generalized to more complex machine learning models. Even in the linear model case, their construction relies on a hyperparameter $c$, and the resulting policy can be highly sensitive to the choice of $c$. Actually, this is common for many pessimism-based RL methods, which often involve some hyperparameter to quantify the degree of pessimism, and the performances rely heavily on the tightness of this uncertainty quantifier. We consider the following toy example to elaborate.

\textbf{A toy example}.
Suppose we model $Q$ via a linear function:
$f(s,a, \wbf) = \wbf^\top \phibf(s,a)$, where $\wbf \in \mathbb{R}^p$ is the coefficient of the linear basis function $\phibf$ and is estimated by a ridge regression following \cite{jin2021pessimism}. They set 
\begin{align}\label{eqn:gamma}
\Gamma(s,a) = cp[\phibf(s,a)^T \Lambda^{-1} \phibf(s,a)]^{1/2}\sqrt{\log (2dn/\xi)},
\end{align}
for some constant $c>0$, where $\Lambda = \sum_{i=1}^n \phi(s_i,a_i) \phi(s_i,a_i)^T + \lambda \Ibf$,  $\lambda$ is the ridge parameter, and $\Ibf$ is the identity matrix. The choice of $c$ in \eqref{eqn:gamma} is crucial for the performance, as a small $c$ would fail to meet the requirement in \ref{eqn:condset} when the data coverage is inadequate, and a large $c$ would over-penalize the Q-function when the coverage is sufficient. Figure \ref{fig:toy} compares the regret of our method and PEVI, where there are two treatments $\{1,2\}$ and a two-dimensional state $S=(S_1,S_2)$. The reward $R$ is generated from a Gaussian distribution with mean $(0.8 + 0.2A)(S_1 + 2S_2)$ and variance $\sigma^2$, and the behavior is generated according to an $\epsilon$-greedy policy that combines a uniformly random policy with a pretrained optimal policy. In this example, $\epsilon$ characterizes the level of the coverage, and we consider two levels $\epsilon = 0.95$ where sub-optimal actions are less explored, and $\epsilon = 0.5$ where the coverage holds. We vary the noise level $\sigma$, and compare our proposed method and PEVI under varying choices of $c = \{0, 1, 2, 5, 10\}$. It is seen that PEVI is highly sensitive to $c$ under different values of $\epsilon$ and $\sigma$. By contrast, our proposed method takes a significance level as the input, which is fixed to $0.9$ or $0.95$ to ensure \eqref{eqn:condset} holds with a large probability, and it achieves a much more stable performance.

\section{BAYESIAN LEARNING WITH PESSIMISM}
\label{sec:meth}

\subsection{Basic Idea: Offline Contextual Bandit} 
\label{subsec:pessbml}

As discussed earlier, the success of the pessimism-based methods relies crucially on the uniform uncertainty quantification of the $Q$-function estimation. Existing solutions require a hyperparameter to properly quantify the degree of pessimism, whereas the choice of such a parameter can be difficult. To address this challenge, and to make the pessimism approach more generally applicable in the offline setting, we propose a data-driven procedure and derive the uniform uncertainty quantification, without  requiring specific models or tuning the degree of pessimism when searching for the optimal decision rules. We first illustrate our idea through a single-stage contextual bandit problem in this section, and discuss the dynamic setting of dynamic treatment regimes in the next section.

Suppose we observe the data $\mathcal{D}_n = \{s_i,a_i,r_i\}_{i=1}^n$. Motivated by Thompson sampling, we propose to model the conditional reward distribution given the state-action pair by $p(r|s,a,w)$, and estimate the model parameter $w\in \mathbb{R}^p$ under a Bayesian framework. Specifically, we first apply BML to obtain the posterior distribution $p(\wbf | \mathcal{D}_n)$, 
and construct a credible set $\mathcal{W}$ given the posterior, such that $P(\wbf \in \mathcal{W} | \mathcal{D}_n) \geq 1 - \alpha$, where $1-\alpha \in (0, 1)$ is the user-specified coverage rate, which usually takes the fixed value of 0.9 or 0.95. Next, instead of choosing an action that maximizes the conditional mean function 
\begin{eqnarray*}
f(s,a,\wbf)=\int_r p(r|s,a,\wbf)dr,
\end{eqnarray*}
with a randomly drawn $\wbf$ as in the online setting, we construct the lower bound of the credible set for $f(s,a,\wbf)$, denoted by $f_L(s, a)$, by solving the following chance constraint optimization problem, 
\begin{align}\label{eqn: chance opt}
\begin{split}
&\underset{\wbf \in \mathcal{W}}{\text{minimize}} \; f(s, a, \wbf), \;\; \\
& \text{ subject to } \;\; \mathbb{P}(\wbf \in \mathcal{W} | \mathcal{D}_n) \geq 1 - \alpha.
\end{split}
\end{align}
Although our credible set is constructed using Bayesian inference, Proposition \ref{prop1} in Section \ref{sec:theory} guarantees that, by the  Bernstein-von Mises theorem, the solution $f_L(s, a)$ to \eqref{eqn: chance opt} provides a valid asymptotic lower bound for $Q(s, a)$ uniformly over $(s, a) \in \mathcal{S} \times \mathcal{A}$ from a frequentist perspective. 

Note that the optimization in \eqref{eqn: chance opt} may not be straightforward. First, it requires to specify the credible set $\mathcal{W}$ that satisfies the coverage constraint. This can be challenging for complex nonlinear models where the exact Bayesian inference is intractable. Second, it can be computationally difficult to optimize the objective function $f(s,a,\wbf)$ with the inequality constraint. 

To address the first challenge, we adopt the variational inference approach, and parameterize the posterior function using a Gaussian distribution $\mathcal{N}(\widehat{\wbf}, \widehat\Sigma)$. The Gaussian model is correctly specified for BLBM, and provides a valid approximation for a large number of nonlinear models \citep{wang2019frequentist}. Under the Gaussian approximation, the posterior distribution of $(\wbf - \widehat{\wbf})^\top \widehat\Sigma^{-1} (\wbf - \widehat{\wbf})$ follows a $\chi^2$ distribution with the degree of freedom $p$, based on which we can easily construct $\mathcal{W}$. 

To address the second challenge, we note that it is relatively straightforward to evaluate the objective function at feasible points. Therefore, we propose a sampling-based algorithm that first randomly collects $N$ samples from the posterior distribution, denoted as $\{\wbf_1, \ldots, \wbf_N\}$. Among these sampling points, we compare the objective values that satisfy the quadratic constraint in \eqref{eqn: quadratic constraint}, and select the smallest one, denoted by $\wbf^\ast$. This yields the following optimization problem,
\begin{align}\label{eqn: quadratic constraint}
\begin{split}
&\underset{j\in \{1,\cdots,N\} }{\text{minimize}} \; f(s,a, \wbf_j), \;\; \\
&\text{ subject to } \;\; (\wbf_j - \widehat{\wbf})^\top \widehat \Sigma^{-1} (\wbf_j - \widehat{\wbf})\leq \chi^2_{1-\alpha}(p),
\end{split}
\end{align}
where $\chi^2_{1-\alpha}(p)$ is the $(1-\alpha)$th quantile of the $\chi^2$ distribution.  

We denote the final solution by $\widehat f^{}_L(s, a) = f(s, a, \wbf^\ast)$. Proposition \ref{prop2} in Section \ref{sec:theory} shows that this solution $\widehat f_L(s, a)$ based on the Gaussian approximation and Monte Carlo sampling provides a valid uniform lower bound for $Q$-function estimation. 

Finally, we output the greedy policy with respect to  $\widehat{f}_L(s, a)$ as $\widehat{\pi}(s) \in \text{argmax}_{a \in \mathcal{A}} \widehat{f}_L(s, a)$ for any $s \in \mathcal{S}$. We summarize our procedure in Algorithm \ref{alg:supervised learning}.

\begin{algorithm}[t!] 
\caption{Pessimism-based Bayesian learning for offline contextual bandit.} \label{alg:supervised learning}
\textbf{Input}: The observed data $\mathcal{D}_n = \{s_i, a_i, r_i\}_{i=1}^n$, and the significance level $\alpha$. \\
\textbf{Step 1}: Fit BLBM or BNN on $\mathcal{D}_n$. \\
\textbf{Step 2}: Compute the posterior distribution $p(\wbf|\mathcal{D}_n)$, with mean $\widehat{\wbf}$ and covariance $\widehat \Sigma$ estimated by BLBM or BNN.\\
\textbf{Step 3}: Draw $N$ random samples $\{\wbf_i \}_{i=1}^N$ from $p(\wbf|\mathcal{D}_n)$, and obtain the index set $J = \{j \in [N] \, \mid \, (\wbf_j - \widehat{\wbf})^\top \widehat \Sigma^{-1} (\wbf_j - \widehat{\wbf})\leq \chi^2_{1-\alpha}(p) \}$.\\
\textbf{Step 4}: Choose $j^\ast = \underset{j \in J}{\text{argmin}} f(s, a,\wbf_j)$. Set $\wbf^\ast =\wbf_{j^\ast}$, and $\widehat{f}_L(s, a) = f(s, a, \wbf^\ast)$.\\
\textbf{Step 5}: Compute the estimated optimal policy as $\widehat{\pi}(s) \in \text{argmax}_{a \in \mathcal{A}} \widehat{f}_L(s, a)$, for any $s \in  \mathcal{S}$.\\
\textbf{Output: } A uniform lower bound $\widehat{f}_L$, and the estimated optimal policy $\widehat \pi$.
\end{algorithm}

\subsection{Dynamic Treatment Regimes}
\label{subsec:dtr}

We next extend our method to the DTR problem, where the insufficient data coverage becomes more serious as the number of decision stages increases.  

Suppose we observe the data $\mathcal{D}_n = \left\{ s^{(1)}_i,a^{(1)}_i,s^{(2)}_i,a^{(2)}_i, \ldots, s^{(T)}_i,a^{(T)}_i, r^{(T)}_i \right\}_{i=1}^n$ consisting of i.i.d.\ realizations of state, action and reward tuples $\{S^{(1)}, A^{(1)}, S^{(2)}, A^{(2)}, \ldots, S^{(T)}, A^{(T)}, R^{(T)}\}$ at $T$ stages. Denote $H^{(t)} = (S^{(1)}, A^{(1)}, \ldots, S^{(t)})$ as the history information up to the decision point $t$, and its realization for each instance as $h^{(t)}_i$ for $i = 1, \ldots, n$. Here we only consider the sparse reward setting commonly seen in medical applications \citep{murphy2003optimal}, but our method can also be applied when there is an immediate reward at each decision point. We propose to incorporate our pessimism-based BML idea at each stage. Specifically, at the last stage, similar as in single-stage contextual bandit, we first construct the uniform lower confidence bound $\widehat{f}^{(T)}_L(h^{(T)}, a^{(T)})$ for $\mathbb{E}(R^{(T)} | H^{(T)} = h^{(T)}, A^{(T)} = a^{(T)})$. We then obtain the estimated optimal policy at the last stage as
\begin{align*}
\widehat{\pi}_T(h^{(T)}) \in \text{argmax}_{a^{(T)} \in \mathcal{A}} \quad \widehat{f}^{(T)}_L(h^{(T)}, a^{(T)}),
\end{align*}
for every $h_T$. Next, to estimate the optimal policy for the $(T-1)$-stage, we employ dynamic programming, and construct the pseudo-reward for each instance at the $(T-1)$-stage as $$r^{(T-1)}_i = \max_{a^{(T)} \in \mathcal{A}} \widehat{f}^{(T)}_L(h^{(T)}_i, a^{(T)}),$$ for $i=1, \ldots, n$. We then apply Algorithm \ref{alg:supervised learning} again, using $\{h^{(T-1)}_i,a^{(T-1)}_i,r_i^{(T-1)}\}_{i=1}^n$ to construct the uniform lower confidence bound $\widehat{f}^{(T-1)}_L(h^{(T-1)}, a^{(T-1)})$ for 
\begin{align*}
\mathbb{E}(\max_{a \in \mathcal{A}}\widehat{f}_L^{(T)}(H^{(T)}, a^{(T)}) | H^{(T-1)} = h^{(T-1)}, A^{(T-1)} = a)
\end{align*}
for every $h^{(T-1)}$ and $a$. We obtain the estimated optimal policy at the $(T-1)$-stage as
\begin{align*}
\widehat{\pi}_{T-1}(h^{(T-1)}) \in \text{argmax}_{a \in \mathcal{A}} \quad \widehat{f}^{(T-1)}_L(h^{(T-1)}, a),
\end{align*}
for every $h^{(T-1)}$. We iterate the above process until the estimated optimal policy of the first stage is obtained. We summarize our proposed procedure in Algorithm \ref{alg:uni_stage2}. We also remark that dynamic treatment regimes differ from MDPs that impose the Markov assumption within each trajectory, in that, in our setting, the Markov assumption can be violated and the optimal treatment regime at each stage depends on the full data history.

\begin{algorithm}[t!] 
\caption{Pessimism-based Bayesian learning for multi-stage dynamic treatment regimes.} \label{alg:uni_stage2}
\textbf{Input:} The observed data $\mathcal{D}_n = \{s^{(1)}_i,a^{(1)}_i,s^{(2)}_i,a^{(2)}_i, \ldots, s^{(T)}_i,a^{(T)}_i, r^{(2)}_i\}_{i=1}^n$, the length of horizon $T$, and the significance levels $\alpha_1, \ldots, \alpha_T$.  \\
\textbf{Initialize} $\widehat f^{T+1}_{L} =0$.\\
\textbf{For} $K = T, T-1, \cdots, 1$\\
\noindent  
 \qquad Apply Algorithm \ref{alg:supervised learning} using the data rearranged as $\{ s_i = h^{(K)}_i, a_i = a^{(K)}_i , r_i = r^{(K)}_i\}_{i =1}^n$, with the confidence level $\alpha_K$ to obtain $\widehat \pi_K$ with the estimated lower confidence bound $\widehat{f}^{(K)}_L$. \\
\textbf{State}: Construct the pseudo-reward, $r^{(K-1)}_i = \max_{a \in \mathcal{A}} \widehat{f}^{(K)}_L(h^{(K)}_i, a)$, for $i = 1, \ldots, n$. \\
\textbf{End for}.\\
\textbf{Output: } The estimated optimal policy $\{\widehat \pi_t \}_{1 \leq t \leq T}$.
\end{algorithm}

\section{THEORY} \label{sec:theory}

We next establish theoretical guarantees for our proposed method.  We focus on the setting of offline contextual bandit of Section \ref{subsec:pessbml} here, and extend the results to the DRT setting in Appendix \ref{appendix_dtr}. 

We first list a set of regularity conditions. Recall that $p(r|s, a, \wbf)$ corresponds to the model we impose for the conditional reward distribution given the state-action pair. 
\begin{assump} \label{assump1}
\begin{enumerate}[(i)]
\item The realization condition holds, i.e., there exists some $\wbf_0$, such that $p(r|s,a,\wbf_0)$ is the oracle conditional reward density function. 
\item The parameter space of $\omega$ is compact, and $p(r|s, a, \wbf)$ is continuous and identifiable in $\omega$.
\item $p(r|s, a, \wbf)$ is differentiable in quadratic mean at the oracle parameter $\wbf_0$ with a non-singular Fisher information matrix.
\item The prior measure of $\wbf$ is absolutely continuous in a neighborhood of $\wbf_0$ with a continuous positive density at $\wbf_0$.
\end{enumerate}
\end{assump}
Assumption \ref{assump1} imposes the conditions on the parameter space and smoothness of the conditional density function, so that we can apply the Bernstein-von Mises theorem, and in turn establish the asymptotic equivalence between the derived credible interval and the confidence interval from the frequentist perspective. These conditions are all mild and standard in the literature  \citep{kleijn2012bernstein,bickel2012semiparametric,kim2006bernstein}.

We next obtain the following proposition. 

\begin{proposition} \label{prop1}
Suppose Assumption \ref{assump1} holds.  
Then,  
\begin{align*}
\liminf_{n \rightarrow \infty} \mathbb{P}\left(\cap_{(s,a)}\left\{f_L(s,a) \leq Q(s, a) \right\} \right) \geq 1 - \alpha.
\end{align*}
\end{proposition}
Note that $f_L(s,a)$ is the theoretical lower bound obtained by solving the exact optimization \eqref{eqn: chance opt} and $Q(s, a) = f(s, a, \wbf_0)$ with the oracle parameter $\wbf_0$ under the realization condition in Assumption \ref{assump1}(i). Proposition \ref{prop1} ensures that solving 
\eqref{eqn: chance opt} is asymptotically equivalent to construct a valid and uniform lower bound, and thus the validity of using the Bayesian learning approach for quantifying the degree of pessimism. 

Since the exact optimization \eqref{eqn: chance opt} is difficult to solve, we next extend the above proposition to the case where Gaussian approximation and Monte Carlo sampling are applied to approximate the lower bound. 
\begin{proposition}  \label{prop2}
Suppose Assumption \ref{assump1} holds. Then, 
\begin{align*}
\liminf_{n \rightarrow \infty}   \liminf_{N \rightarrow \infty}\mathbb{P}\left( \cap_{(s,a)} \left\{ \widehat f_L(s, a) \leq Q(s,a)\right\}   \right) \geq 1 - \alpha.
\end{align*}
\end{proposition}
Note that $\widehat f_L(s, a)$ is the lower bound obtained by solving the surrogate optimization \eqref{eqn: quadratic constraint}. Proposition \ref{prop2} ensures that the resulting solution based on the Gaussian approximation and Monte Carlo sampling provides a valid uniform lower bound asymptotically.

Finally, we establish the theoretical guarantee that characterizes the average regret of the estimated optimal policy from Algorithm \ref{alg:supervised learning}, i.e., the difference between the value function under the optimal policy and that under the estimated policy. 

\begin{theorem}\label{thm2}
Suppose Assumption \ref{assump1} holds. Then, as $n,N\to \infty$, with probability at least $1-\alpha+o(1)$, the average regret is upper bounded by
\begin{align*}
\mathbb{E}_{\pi^*}\left[\mathbb{E}(R | S, A) - \widehat{f}_L(S,A)\right].  
\end{align*}
Specifically, if we use BLBM with $p$ basis functions for model fitting, then there exists some constant $\bar{c}>0$, such that the average regret can be upper bounded by 
\begin{align*}
\bar{c} p^{1/2} n^{-1/2}.
\end{align*}
\end{theorem}
We note that the expectation $\mathbb{E}_{\pi^*}$ in the regret bound is taken with respect to the optimal policy $\pi^*$. In addition, the difference within the square brackets measures the estimation error of the Q-function. As such, the average regret of the proposed policy depends only on the estimation error of the optimal action's Q-estimator, instead of the uniform estimation error of the Q-estimator at each action. The latter can be much larger without the full coverage assumption.  In the case of BLBM, $N$ is not included in the upper bound since we can explicitly solve optimization (\ref{eqn: quadratic constraint}) without Monte Carlo sampling. 

We also remark that our theory can be extended to obtain finite-sample guarantee as well. As an example, \cite{hipp1976bernstein} showed that, under some regularity conditions, the Bernstein-von Mises approximation of the posterior distribution was of the order $n^{-1/2}$. Following similar arguments, we can further extend Theorem 1 to obtain a nonasymptotic probability bound. We provide more details in Corollary 1 in Appendix \ref{appendix_cor1}.

\section{SYNTHETIC DATA ANALYSIS}
\label{sec:exper}

\textbf{Simulation Setup}. We conduct extensive numerical experiments to investigate the empirical performance of the proposed pessimism-based Bayesian learning method (PBL). We illustrate our method using both BLBM and BNN. We also compare with the method of \citet[PEVI]{jin2021pessimism}, and a standard Q-learning method using BLBM or BNN but without pessimism (Non-Pessi). 

We consider both a single-stage contextual bandit setting and a two-stage DTR setting. In the two-stage setting, since a linear Q-function model is likely to be misspecified in backward induction, we did not implement BLBM under this setting. For both settings, we consider two data generating processes, with linear and nonlinear Friedman signals \citep[see e.g.,][]{zhao2017selective}, respectively. We generate all actions by the $\epsilon$-greedy policy, with $\epsilon\in \{0.95,0.85,0.75,0.5\}$, and a smaller $\epsilon$ indicating a larger coverage over the state-action distribution. We choose the sample size $n$ from $\{500, 1000, 1500, 2000, 2500, 3000\}$, and repeat each experiment $50$ times. More details of the data generation and implementations are given in the Appendices \ref{appendix_sim1}, \ref{appendix_sim2} and \ref{appendix_ctime}. To implement PEVI, we choose the hyperparameter $c$ from $\{1,2,5,10\}$. We also conduct a sensitivity analysis by varying a number of parameters in Appendix \ref{appendix_sensi}, including the number of Monte Carlo samples $N$, the number of ensembles $M$ for the MC gradient computation in variational inference, and the significance level $\alpha$. We find that our method is not overly sensitive to those parameters as long as they are in a reasonable range. We make our code publicly available at \url{https://github.com/yunzhe-zhou/PBL}.

\begin{figure*}[t!]
\begin{subfigure}{1\textwidth}
\centering
\includegraphics[width=0.9\textwidth,height=2.35in]{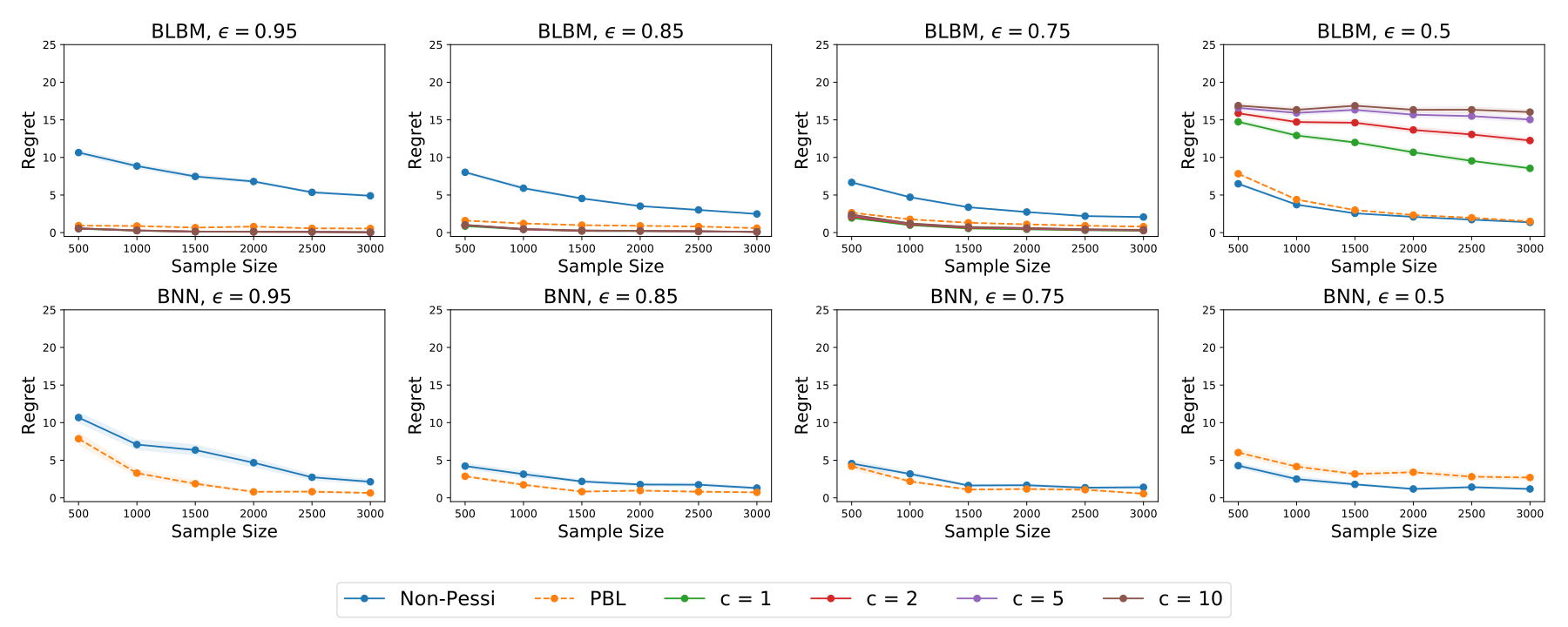}
\caption{Linear Signal}
\label{stage1_linear}
\end{subfigure}
\begin{subfigure}{1\textwidth}
\centering
\includegraphics[width=0.9\textwidth,height=2.35in]{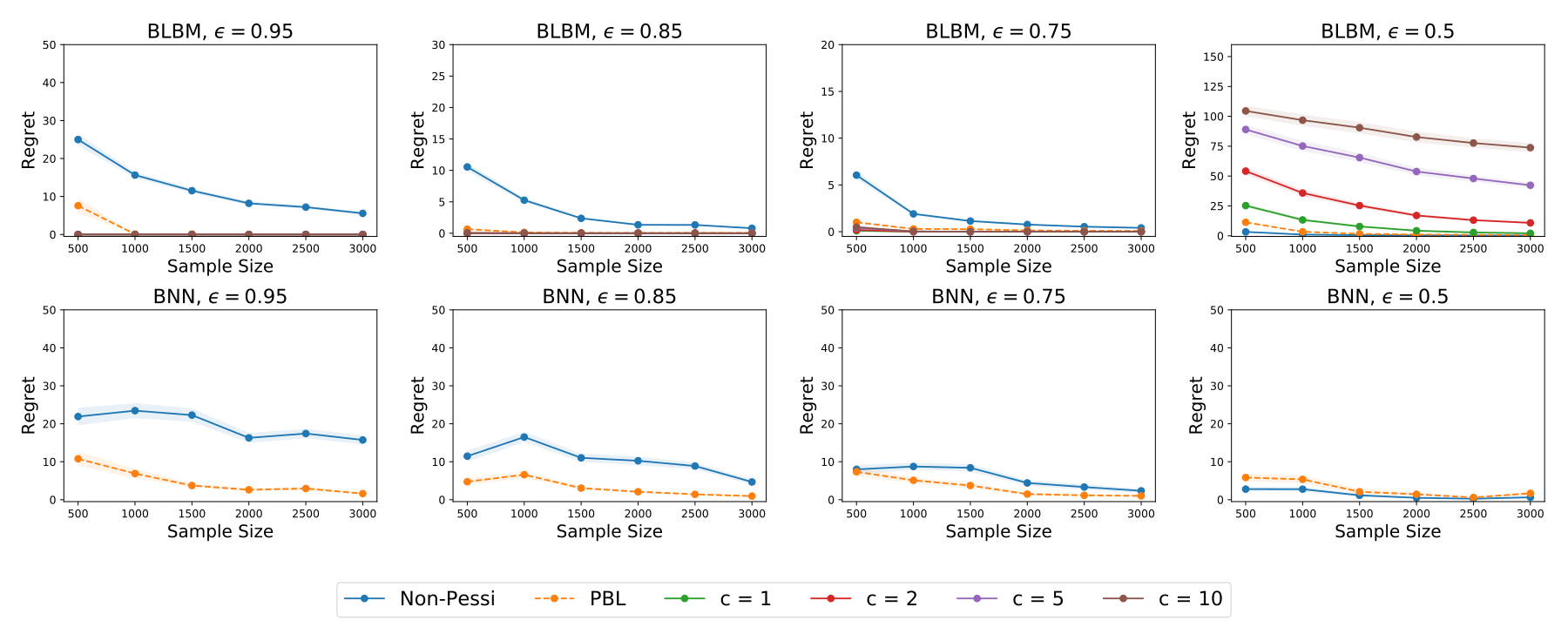}
\caption{Nonlinear Signal}
\label{stage1_nonlinear}
\end{subfigure}
\caption{The single-stage contextual bandit simulation with linear and nonlinear signals. The methods compared include the proposed PBL method, the PEVI method of \cite{jin2021pessimism}, and the standard Q-learning method without pessimism, each of which using BLBM and BNN.}
\label{fig:stage1}
\end{figure*}

\textbf{Results}. Figure \ref{fig:stage1} reports the results for the single-stage contextual bandit setting under the linear and nonlinear signals, whereas Figure \ref{fig:stage2} in Appendix \ref{appendix_plot} reports the results for the two-stage DTR setting. It is clearly seen from these plots that both our proposed PBL and the PEVI method outperform the standard Q-learning method when $\epsilon\ge 0.75$ and the coverage assumption is seriously violated, demonstrating the advantage of the pessimism principle. Nevertheless, for the single-stage setting when $\epsilon=0.5$ and the coverage is of less concern, PEVI over-penalizes the Q-function, leading to a large regret. By contrast, our proposed method performs comparably to the standard Q-learning algorithm in this setting. For the two-stage setting, our proposed method based on BNN outperforms PEVI in all cases. This is because PEVI uses a linear function approximation. The linearity assumption is likely to be violated in backward induction, leading to sub-optimal policies.

\section{REAL DATA APPLICATION}
\label{sec:real}

We illustrate our method with the MIMIC-III v1.4 dataset that contains critical care data for over 40,000 patients from the Beth Israel Deaconess Medical Center between 2001 and 2012 \citep{johnson2016mimic}. Following the analysis of \cite{raghu2017deep}, we define a $5 \times 5$ action space by discretizing both medical interventions intravenous (IV) fluid and maximum vasopressor (VP) dosage into 5 levels. We define the reward as the negative value of the SOFA score that measures the organ failure of the patients \citep{lambden2019sofa}, so a larger reward is better. We consider the state space with 47 physiological features, including the demographics, lab values, vital signs, and intake and output events. We construct two datasets, one for single–stage contextual bandit, and the other for two-stage DTR. We randomly split each data into a training set and a testing set with equal sample size. We apply our algorithm to the training data and use BNN to fit the Q-function. We did not use BLBM or apply PEVI to this data, since the associations between the features and rewards are expected to be highly complex and nonlinear \citep{raghu2017deep}. We compare our proposed PBL method with the standard Q-learning method based on BNN without using pessimism, as well as the conservative Q-learning (CQL) method implemented via the \textsf{d3rlpy} package at its default setting. 

\begin{figure}[h]
\begin{subfigure}{0.5\textwidth}
\includegraphics[width=0.95\textwidth]{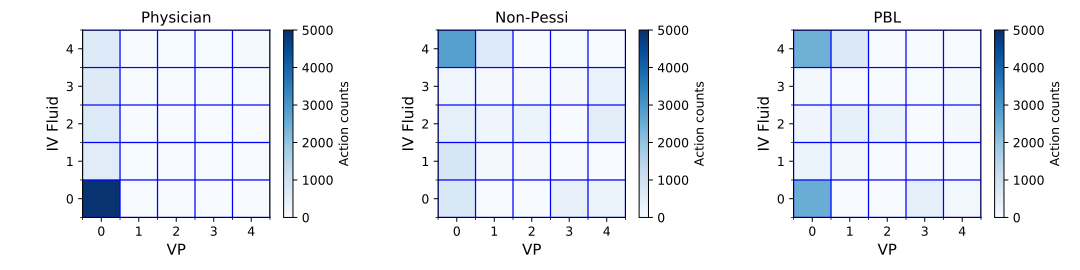}
\caption{Single–stage decision making}
\label{fig:sub1}
\end{subfigure}
\begin{subfigure}{0.5\textwidth}
\includegraphics[width=0.95\textwidth]{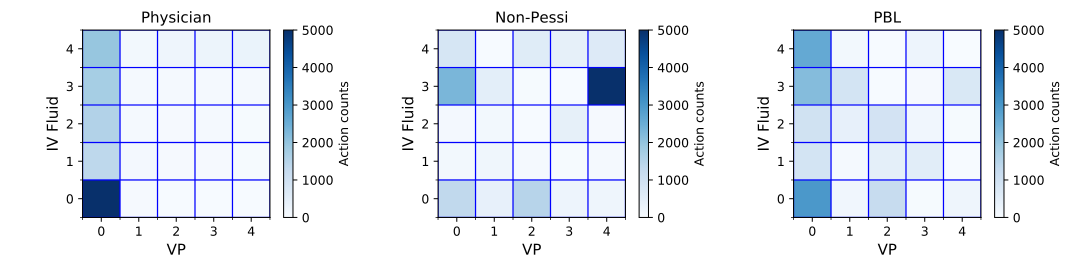}
\caption{Two-stage decision making}
\label{fig:sub2}
\end{subfigure}
\caption{Heatmap for the actions generated by the physician policy and the learned policy by the proposed PBL method and the standard Q-learning method without pessimism.}
\label{fig:real}
\end{figure}

\begin{figure}[h]
\centering
\includegraphics[width=0.4\textwidth,height=4cm]{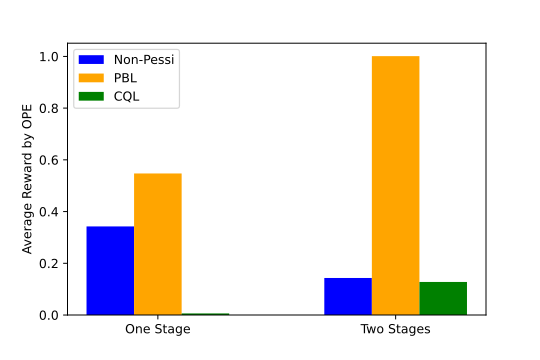}
\caption{Average reward of proposed PBL method, the standard Q-learning method, and the conservative Q-learning evaluated by OPE.}
\label{fig:real_reward}
\end{figure}

Figure \ref{fig:real} reports the frequencies of the assigned treatments, in terms of heatmaps, given by the physicians, the proposed PBL, and the standard Q-learning method (Non-Pessi). For each heatmap, the axis labels show different levels of each action, where $0$ represents no drug is given, and a nonzero value corresponds to the dosage of the IV fluid or VP. It can be seen from the plot that the physicians tended to prescribe no vasopressor to patients, but often considered IV fluids with various dosages. Meanwhile, the policy produced by our pessimism-based method tends to recommend treatment $0$ or $4$ for IV fluids and treatment $0$ for VP, which is consistent with physicians' recommendations to some extent. Moreover, we use 5-fold cross-validation and apply the importance sampling method \citep{zhang2013robust} to evaluate the average reward under the estimated optimal policies produced by the proposed PBL, the standard Q-learning and CQL. Figure \ref{fig:real_reward} reports the results. It is seen that PBL achieves the highest average award among the three methods, demonstrating the competitive performance of our method in this real data application.

\section{DISCUSSIONS}

In this article, we develop a novel pessimism-based Bayesian learning approach for offline optimal dynamic treatment regimes. We propose to combine the pessimism principle with Thompson sampling and Bayesian machine learning to optimize the degree of pessimism. Theoretically, we derive the upper bound for the regret of the proposed method, and obtain its explicit form in a specific case of a parametric model. Empirically, we develop a highly efficient and scalable computational algorithm based on variational inference. We also conduct extensive numerical experiments to illustrate the superior performance of our method. In terms of potential limitations of our proposed method, since it requires a large number of Monte Carlo samples, it can be computationally intensive, especially when the model dimension is large. How to further improve the computational efficiency warrants future research. In addition, it is of interest to extend our theoretical results that take the model misspecification and approximation errors into consideration, and we leave it as future research.

\subsubsection*{Acknowledgements}
Shi's research was partly supported by the EPSRC grant EP/W014971/1. Li’s research was partly supported by the NSF grant CIF-2102227, and the NIH grants R01AG061303 and R01AG062542. The authors thank all the constructive comments from the referees and the area chair, which have led to a significant improvement of the earlier version of this paper.

\bibliography{main}

\appendix

\onecolumn
\renewcommand{\thefigure}{A\arabic{figure}}
\setcounter{figure}{0}
\renewcommand{\thetheorem}{A\arabic{theorem}}
\setcounter{theorem}{0}
\renewcommand{\theproposition}{A\arabic{proposition}}
\setcounter{proposition}{0}
\renewcommand{\theassump}{A\arabic{assump}}
\setcounter{assump}{0}

\section{ADDITIONAL THEORETICAL RESULTS}
\label{appendix_theory}

\subsection{Theory for Dynamic Treatment Regimes}
\label{appendix_dtr}

We first generalize our theoretical guarantee for the DTR setting. We introduce the following notation. For any $K = 1,2,\ldots,T-1$, let $f^{(K)}(h_K, a_K, \wbf)$ denote the model at stage $K$ used to fit the response of pseudo-reward $\max_{a^{(K+1)}} \widehat{f}^{(K+1)}_L(h^{(K+1)}, a^{(K+1)})$. For $K = T$, let $f^{(T)}(h_T, a_T, \wbf)$ denote the model at final stage $T$ for fitting  $\mathbb{E}(R^{(T)} | H^{(T)} = h^{(T)}, A^{(T)} = a^{(T)})$. Denote  $p^{(K)}(r|h_K, a_K, \wbf)$ as the conditional density of the pseudo-reward given the history information under the model $f^{(K)}(h_K, a_K, \wbf)$, such that
\begin{eqnarray*}
f^{(K)}(h_K, a_K, \wbf)=\int_r p^{(K)}(r|h_K, a_K, \wbf)dr,
\end{eqnarray*}

We consider the following assumption for $K=1,2,\ldots,T$:
\begin{assump} \label{assump_appendix1}
\begin{enumerate}[(i)]
\item The realization condition holds, i.e., there exists some $\wbf^{({K})}_0$, such that $p^{(K)}(r|h_K,a_K,\wbf^{({K})}_0)$ is the oracle conditional pseudo-reward density function. 
\item The parameter space of $\omega$ is compact, and $p^{(K)}(r|h_K,a_K,\wbf)$ is continuous and identifiable in $\omega$.
\item $p^{(K)}(r|h_K,a_K, \wbf)$ is differentiable in quadratic mean at the oracle parameter $\wbf^{({K})}_0$ with a non-singular Fisher information matrix.
\item The prior measure of $\wbf$ is absolutely continuous in a neighborhood of $\wbf^{({K})}_0$ with a continuous positive density at $\wbf^{({K})}_0$.
\end{enumerate}
\end{assump}

\begin{assump} \label{assump_appendix3}
Suppose the data used for fitting $\widehat{f}^{({K})}$ for $K=1, \ldots T$ are independent. 
\end{assump}
Assumption \ref{assump_appendix1} is similar as Assumption \ref{assump1}, but extends to multiple stages. Assumption \ref{assump_appendix3} imposes the cross-fitting condition to simplify our theoretical analysis. Without such an independence assumption, we need to impose certain entropy condition on the function class to prove Theorem \ref{thm_appendix} \citep{vaart1996weak}.

We next obtain the following proposition, and show that the $\widehat f^{(K)}_L(h_K, a_K)$ based on the Gaussian approximation and Monte Carlo sampling provides a valid uniform lower bound from the frequentist perspective.

\begin{proposition} \label{prop_appendix1}
Suppose Assumptions \ref{assump_appendix1} and \ref{assump_appendix3} hold. Then,  
\begin{align*}
\liminf_{\substack{n \rightarrow \infty \\ N \rightarrow \infty}}   \mathbb{P}\left( \cap_{(h_K,a_K)} \left\{ \widehat f^{(K)}_L(h_K, a_K) \leq Q^{(K)}(h_K,a_K)\right\}   \right) \geq 1 - \alpha.
\end{align*}
\end{proposition}

Finally, we establish the theoretical guarantee that characterizes the average regret of the estimated optimal policy from Algorithm \ref{alg:uni_stage2}.

\begin{theorem}\label{thm_appendix}
Suppose Assumptions \ref{assump_appendix1} and \ref{assump_appendix3} sequentially hold for each stage $K$. Suppose the significance level $\alpha_K=\alpha/T$ is set following the Bonferroni correction. Then, as $n,N\to \infty$, with probability at least $1-\alpha+o(1)$, the average regret is upper bounded by
\begin{align*}
\sum_{K=1}^T \mathbb{E}_{\pi^*}\left[\mathbb{E}(R_K | H_K, A_K) - \widehat{f}^{(K)}_L(H_K, A_K)\right].  
\end{align*}
Specifically, if we use BLBM with $p$ basis functions for model fitting, then there exists some constant $\bar{c}>0$, such that the average regret can be upper bounded by 
\begin{align*}
\bar{c} T p^{1/2} n^{-1/2}.
\end{align*}
\end{theorem}

\subsection{Finite-sample Guarantee for Contextual Bandit} 
\label{appendix_cor1}

We next extend our theory to obtain the finite-sample guarantee for contextual bandit. As an example, following similar arguments in analyzing the Bernstein-von Mises approximation \citep{hipp1976bernstein}, we can show that the statement of Theorem 1 holds with probability at least $1-\alpha-C n^{-1/2}$, for some positive constant $C$ that depends on the number of parameters $p$ only. This yields a non-asymptotic probability upper bound as follows. 

\begin{corollary}
Suppose Assumption \ref{assump1} holds. Then, as $N\to \infty$, with probability at least $1-\alpha-C n^{-1/2}$, the average regret is upper bounded by
\begin{align*}
\mathbb{E}_{\pi^*}\left[\mathbb{E}(R | S, A) - \widehat{f}_L(S,A)\right].  
\end{align*}
\end{corollary}
where $C$ is a positive constant that depends on the number of parameters $p$.

\section{PROOFS} \label{proof}

We provide the proofs for Proposition \ref{prop_appendix1} and Theorem \ref{thm_appendix} in Appendix \ref{appendix_dtr}. By setting $T=1$ as a special case, we obtain the proofs for Proposition \ref{prop2} and Theorem \ref{thm2} in Section \ref{sec:theory}.

\subsection{Proof of Proposition \ref{prop_appendix1}}
Denote $ \mathcal{H}$ as the space for the history information. By definition that $f^{(K)}(h_K, a_K, \wbf)$ is the model at stage $K$ used to fit the response $\max_{a^{(K+1)}} \widehat{f}^{(K+1)}_L(h^{(K+1)}, a^{(K+1)})$, for any $K = 1,2,\ldots,T-1$, and by \eqref{eqn: chance opt}, we know that, for any $(h_K,a_K) \in \mathcal{H} \times \mathcal{A}$ and $\wbf\in \mathcal{W}$,
\begin{align*}
\mathbb{P}\left(\left\{ \wbf: \forall (h_K,a_K) \in  \mathcal{H} \times \mathcal{A}, f^{(K)}_L(h_K,a_K) \leq f^{(K)}(h_K, a_K,\wbf)\right\} \, \mid \, \mathcal{D}_n  \right) \geq \mathbb{P}(\wbf \in \mathcal{W} \mid \mathcal{D}_n) \geq 1 - \alpha\nonumber.
\end{align*}
For  $Q^{(K)}(h_K,a_K) = f^{(K)}(h_K, a_K,\wbf^{({K})}_0)$, we obtain that
\begin{align*} 
\mathbb{P}\left(\cap_{(h_K,a_K)}\left\{f^{(K)}_L(h_K,a_K) \leq Q^{(K)}(h_K, a_K) \right\} \, \mid \, \mathcal{D}_n \right)\geq 1 - \alpha\nonumber.
\end{align*}

Recall the assumptions that the parameter space is compact, and the likelihood function $p^{(K)}(r|h_K, a_K, \wbf)$ is continuous and identifiable in $\wbf$. Furthermore, recall that $p^{(K)}(r|h_K, a_K, \wbf)$ is differentiable in quadratic mean at $\wbf^{(K)}_0$ with non-singular Fisher information matrix, and the prior measure of $\wbf$ is absolutely continuous in a neighborhood of $\wbf^{(K)}_0$ with a continuous positive density at $\wbf^{(K)}_0$. Then, by Corollary 7 of \cite{wang2019frequentist}, we have that
\begin{align*}
\liminf_{n \rightarrow \infty} \mathbb{P}\left(\cap_{(h_K,a_K)}\left\{f^{(K)}_L(h_K,a_K) \leq Q^{(K)}(h_K, a_K) \right\} \right) \geq 1 - \alpha.
\end{align*}

Denote $\wbf_{\text{opt}} = \underset{\wbf \in \mathcal{W}}{\text{minimize}} \; f^{(K)}(h_K, a_K, \wbf)$, subject to $\mathbb{P}(\wbf \in \mathcal{W} | \mathcal{D}_n) \geq 1 - \alpha$, which is the solution to the optimization problem in \eqref{eqn: chance opt} for a given $a_K$ and $h_K$. Let $\delta > 0$ be a positive constant, such that $|\wbf_{\text{opt},j} - \wbf_{j}| \leq \delta$ for $j \in \{1,2,\ldots,p\}$, and the value of $\delta$ will be determined later. Since $f$ is Lipschitz continuous in $\wbf$, there exists a constant $L>0$, such that, for $ \forall (h_K,a_K) \in \mathcal{H} \times \mathcal{A}$,
\begin{align*}
|f^{(K)}(h_K,a_K,\wbf_{\text{opt}}) - f^{(K)}(h_K,a_K,\wbf)|  \leq L||\wbf_{\text{opt}} - \wbf||_2 \leq L\sqrt{p}\delta.
\end{align*}
Considering the interval $I_j = [\wbf_{\text{opt},j} -\delta,\wbf_{\text{opt},j} + \delta]$, we have that
\begin{align*}
\mathbb{P}\left(\wbf_j \in I_j \, \mid \, \mathcal{D}_n  \right) = \int_{I_j} p(\wbf_j|\mathcal{D}_n) d \wbf_j.
\end{align*}
In our method, we adopt a Gaussian distribution to approximate the posterior distribution. Hence, $p$ is Lipschitz continuous in $\wbf_j$ for a given mean and a given covariance matrix. Thus, we can find some constants $c_1,c_2>0$, such that 
\begin{align*}
c_1 \delta \leq   \mathbb{P}\left(\wbf_j \in I_j \, \mid \, \mathcal{D}_n  \right) \leq c_2 \delta.
\end{align*}
This implies that
\begin{align*}
\mathbb{P} \left(\bigcap_{h_K,a_K} \left\{|f^{(K)}(h_K,a_K,\wbf_{\text{opt}}) - f^{(K)}(h_K,a_K,\wbf_j)|  \leq  L\sqrt{p}\delta\right\} \, \mid \, \mathcal{D}_n  \right)  \geq (c_1 \delta)^p .
\end{align*}
Since we randomly generate $N$ samples from the posterior distribution of $\wbf$ and select the one that minimizes $f$, with some calculations, we have that
\begin{align*}
\mathbb{P} \left(|f^{(K)}(h_K,a_K,\wbf_{\text{opt}}) - f^{(K)}(h_K,a_K,\wbf^*)|  \leq  L\sqrt{p}\delta\, \mid \, \mathcal{D}_n  \right) \geq 1-[1 - (c_1 \delta)^p]^N.
\end{align*}
Letting $\delta = N^{-1/(2p)}$, we obtain that
\begin{align} \label{asym_N}
\liminf_{N \rightarrow \infty} \mathbb{P} \left(|f^{(K)}(h_K,a_K,\wbf_{\text{opt}}) - f^{(K)}(h_K,a_K,\wbf^*)| \leq \epsilon \, \mid \, \mathcal{D}_n  \right) = 1
\end{align}
for any $\epsilon$. From Proposition \ref{prop1}, we know that
\begin{align*}
\mathbb{P}\left(\cap_{(h_K,a_K)}\left\{f^{(K)}(h_K,a_K,\wbf_{\text{opt}}) \leq Q^{(K)}(h_K,a_K)\right\} \, \mid \, \mathcal{D}_n  \right) \geq 1 - \alpha.
\end{align*}
Combining it with \eqref{asym_N}, we obtain that, for any $\epsilon$,
\begin{align*}
\liminf_{N \rightarrow \infty}\mathbb{P}\left(\cap_{(h_K,a_K)} \left\{\widehat f^{(K)}_L(h_K, a_K) \leq Q^{(K)}(h_K,a_K) + \epsilon\right\} \, \mid \, \mathcal{D}_n  \right) \geq 1 - \alpha,
\end{align*}
where $\widehat{f}_L(h_K, a_K) = f^{(K)}(h_K,a_K,\wbf^*)$. Since $\epsilon$ can be chosen arbitrarily small, we have that 
\begin{align*}
\liminf_{N \rightarrow \infty}\mathbb{P}\left(\cap_{(h_K,a_K)} \left\{\widehat f^{(K)}_L(h_K, a_K) \leq Q^{(K)}(h_K,a_K)\right\} \, \mid \, \mathcal{D}_n  \right) \geq 1 - \alpha,
\end{align*}
By Corollary 7 of \cite{wang2019frequentist}, we have
\begin{align*}
\liminf_{n \rightarrow \infty}  \liminf_{N \rightarrow \infty}\mathbb{P}\left(\cap_{(h_K,a_K)} \left\{\widehat f^{(K)}_L(h_K, a_K) \leq Q^{(K)}(h_K,a_K)\right\} \right)  \geq 1 - \alpha.
\end{align*}

\subsection{Proof of Theorem \ref{thm_appendix}}

In Algorithm \ref{alg:uni_stage2}, we employ dynamic programming and construct the pseudo-reward, $R^{(K)}_i = \max_{a_{K+1} \in \mathcal{A}} \widehat{f}^{({K+1})}_L(h^{({K+1})}_i, a_{K+1})$, for each instance at the $K$th stage, $K=1,2,\ldots,T-1$. Define the event 
\begin{align*}
\Omega_K = \left\{\widehat{f}^{(K)}_L(h_K, a_K) < \mathbb{E}(R_K | H_K = h_K, A_K = a_K), \;\forall (h_K,a_K) \in  \mathcal{H} \times \mathcal{A}\right\}.  
\end{align*}
for $K = 1,2,\ldots,T$. Define the joint event as $\Omega =\bigcap_{K=1}^T \Omega_K$. 

By Proposition \ref{prop_appendix1}, we can show that $\mathbb{P}(\Omega_K) \geq 1-\alpha/T +o(1)$ as both $n$ and $N$ approach infinity for any $K=1,2,\ldots,T$. Then with the Bonferroni correction, we have that $\mathbb{P}(\Omega) \geq 1-\alpha +o(1)$ when $n$ and $N$ approach infinity. Let $\mathbb{E}_T(\mathcal{R};s^{(1)})$ denote the average regret given the initial state $s^{(1)}$. Then we decompose the average regret into three components as follows:
\begin{align*} 
\mathbb{E}_T(\mathcal{R};s^{(1)}) &= \sum_{K=1}^T \Big\{ \underbrace{-  \mathbb{E}_{\hat{\pi}}[\eta_K(h_K,a_K)|H_1 = s^{(1)}]}_{\text{(i)}} + \underbrace{ \mathbb{E}_{\pi^*}[\eta_K(h_K,a_K)|H_1 = s^{(1)}]}_{\text{(ii)}} 
 + \\
&\underbrace{ \mathbb{E}_{\pi^*}[\langle \widehat{f}_L(h_K,a_K),\pi^*(\cdot|h_K)-\hat{\pi}(\cdot|h_K)\rangle_{\mathcal{A}}|H_1 = s^{(1)}]}_{\text{(iii)}} \Big\}
\end{align*}
where $\eta_K(h_K,a_K) = \mathbb{E}(R_K | H_K = h_K, A_K = a_K) - \widehat{f}^{(K)}_L(h_K, a_K)$  is the model evaluation error. 

We start with the last stage $K=T$ and do backward induction. We first note that, since $\hat{\pi}$ is greedy with respect to $\widehat{f}^{(T)}_L(h, a)$, the optimization error (iii) is non-positive. So it can be directly removed from the bound. Next, we consider the error term (i).  Under event $\Omega_K$, we have that 
\begin{align*}
0 \leq  \eta(h_T,a_T) = \mathbb{E}(R_T | H_T = h_T, A_T = a_T) - \widehat{f}^{(T)}_L(h_T, a_T)
\end{align*}
Thus, we obtain that 
\begin{align*}
-  \mathbb{E}_{\hat{\pi}}[\eta(h_T,a_T)|H_1 = s^{(1)}] \leq 0
\end{align*}
We repeat the same produce for $K=T-1,\ldots,1$, and under each event $\Omega_K$, we obtain that
\begin{align*}
 -  \mathbb{E}_{\hat{\pi}}[\eta(h_K,a_K)|H_1 = s^{(1)}] \leq 0 \\
 \mathbb{E}_{\pi^*}[\langle \widehat{f}_L(h_K,a_K),\pi^*(\cdot|h_K)-\hat{\pi}(\cdot|h_K)\rangle_{\mathcal{A}}|H_1 = s^{(1)}] \leq 0 
\end{align*}

Combining the inequalities above, we have that, under the event $\Omega$,
\begin{align*}
-  \sum_{K=1}^T \mathbb{E}_{\hat{\pi}}[\eta(h_K,a_K)|H_1 = s^{(1)}] \leq 0 \\
\sum_{K=1}^T \mathbb{E}_{\pi^*}[\langle \widehat{f}_L(h_K,a_K),\pi^*(\cdot|h_K)-\hat{\pi}(\cdot|h_K)\rangle_{\mathcal{A}}|H_1 = s^{(1)}] \leq 0 
\end{align*}
which implies that 
\begin{align*}
\mathbb{E}_T(\mathcal{R};s^{(1)})  \leq      \sum_{K=1}^T \mathbb{E}_{\pi^*}\left[\mathbb{E}(R_K | H_K = h_K, A_K = a_K) - \widehat{f}^{(K)}_L(h_K, a_K)|H_1 = s^{(1)}\right].
\end{align*}
Taking the integral over the randomness of $s^{(1)}$ on both sides, as $n,N\to \infty$, with probability at least $1-\alpha+o(1)$, we can upper bound the average regret by
\begin{align*}
\sum_{K=1}^T \mathbb{E}_{\pi^*}\left[\mathbb{E}(R_K | H_K, A_K) - \widehat{f}^{(K)}_L(H_K, A_K)\right].  
\end{align*}

For the specific case of BLBM, we have that 
\begin{align*}
\mathbb{E}_T(\mathcal{R};s^{(1)}) &\leq \sum_{K=1}^T \mathbb{E}_{\pi^*}\left[\mathbb{E}(R_K | H_K, A_K) - \widehat{f}^{(K)}_L(H_K, A_K)\right] \\ &\leq \sum_{K=1}^T \mathbb{E}_{\pi^*}\left[\mathbb{E}(R_K | H_K, A_K)  - f^{(K)}_L(H_K, A_K) + f^{(K)}_L(H_K, A_K) - \widehat{f}^{(K)}_L(H_K, A_K)\right] \\
&\leq \underbrace{\sum_{K=1}^T \mathbb{E}_{\pi^*}\left[\mathbb{E}(R_K | H_K, A_K)  - f^{(K)}_L(H_K, A_K) \right]}_{\text{(I)}} + \underbrace{\sum_{K=1}^T \mathbb{E}_{\pi^*}\left[f^{(K)}_L(H_K, A_K) - \widehat{f}^{(K)}_L(H_K, A_K)\right]}_{\text{(II)}}
\end{align*}
where (II) is 0 if we directly solve the optimization problem \eqref{eqn: quadratic constraint} without Monte Carlo sampling. By Corollary 7 of \cite{wang2019frequentist} and the Lipschitz continuous condition for $f^{(K)}$, we have that 
\begin{align*}
\text{(I)}  &\leq \zeta_{1-\alpha} T \mathbb{E}_{\pi^*} \left[ \phi^T(H_K,A_K) \Lambda^{-1}_K \phi(H_K,A_K)\right] + o(n^{-1/2})\\
&\leq \zeta_{1-\alpha} T \sqrt{\sum_{j=1}^p \frac{1}{cn}} + o(n^{-1/2})\\
&\leq c'\zeta_{1-\alpha} T p^{1/2} n^{-1/2},
\end{align*}
where $\zeta_{1-\alpha}$ is the $1-\alpha$ percentile of the standard normal distribution,  and $\Lambda^{-1}_K = \sum_{i=1}^n \phi(H_{K,i},A_{K,i})\phi^T(H_{K,i},A_{K,i})$.

Therefore, we obtain that 
\begin{align*}
\mathbb{E}_T(\mathcal{R};s^{(1)}) \leq  \bar{c} T p^{1/2} n^{-1/2}.   
\end{align*}
for some constant $\bar{c}$.

\section{ADDITIONAL NUMERICAL RESULTS} 
\label{appendix_sim}

\subsection{Data Generation for One-Stage Contextual Bandit}
\label{appendix_sim1}

We first outline the details of our data generation for one-stage contextual bandit. 
\begin{itemize}
    \item Linear Signal: 
    \begin{align*} 
  r(s,a) = \left\{ 
         \begin{array}{lr}
          0.2s_1 + 0.25s_2 + 0.3s_3 + 0.1z \quad \text{if} \; a=1\\
         0.25s_1 + 0.3s_2 + 0.35s_3  + 0.1z \quad   \text{if} \; a=2
         \end{array}
    \right. 
    \end{align*}
    where $z \sim \text{Normal}(0,1)$. We draw $s \in \mathbb{R}^3$ with $s_i \sim \text{Normal}(0,1)$ for $i=1,2,3$. For each state $s$, we denote $a^* = \arg\max\limits_a \mathbb{E}[r(a,s)]$, and generate $a$ with the probability $P(a=a^*) = 1-\epsilon$, where $\mathbb{E}$ is taken with respect to the randomness of the reward function. 
    
    \item Nonlinear Signal:
    We define two transformation functions,
    \begin{align*}
    f_1(s) &=  [0.1\exp(4s_1) + 4/(1+\exp(-20(s_2-0.5)))+3s_3+2s_4+s_5]/2.5,  \\ 
    f_2(s) &=  [0.12\exp(4s_1) + 4.8/(1+\exp(-20(s_2-0.5)))+3.6s_3+2.4s_4+1.2s_5]/2.5, \\
    r(s,a) &= \left\{
         \begin{array}{lr}
             f_1(s) + 0.1z \quad \text{if} \; a=1 \\  
          f_2(s)+ 0.1z \quad \text{if} \; a=2
         \end{array},
    \right. 
  \end{align*}
  where $z \sim \text{Normal}(0,1)$. We draw $s \in \mathcal{R}^5$ with $s_i \sim \text{Uniform}[0,1]$ for $i=1,2,3,4,5$. We generate the actions in the same way as for the  linear signal.
\end{itemize}

\subsection{Data Generation for Two-Stage DTR}
\label{appendix_sim2}

We next outline the details of our data generation for two-stage DTR. 
\begin{itemize}
    \item Linear Signal: We define two transformation functions,
    \begin{align*}
    f_1(s) &= 0.2s_1 + 0.25s_2 + 0.3s_3, \\ 
    f_2(s) &= 0.25s_1 + 0.3s_2 + 0.35s_3.
    \end{align*}
    We first randomly generate the coefficient matrix $W_1 \in \mathbb{R}^{2 \times 3}$ with each entry independently drawn from $\text{Normal}(0,1)$. We then define $W_2 = W_1 + 0.05$, where the sum calculation is element-wise. We fix $W_1$ and $W_2$. For each replication, we draw the state at the first stage as $s^{(1)}=(s^{(1)}_1,s^{(1)}_2)^T \in \mathbb{R}^{2 \times 1}$ with $s^{(1)}_i \sim \text{Normal}(0,1)$ for $i=1,2$. Suppose that the action of the first stage is chosen as $a^{(1)}$, then we generate the state at the second stage as 
    \begin{align*} 
  \left\{ 
         \begin{array}{lr}
         s^{(2)} =  W_1^{T} s^{(1)} + z \quad \text{if} \; a^{(1)}=1\\
        s^{(2)} =  W_2^{T} s^{(1)} + z \quad \text{if} \; a^{(1)}=2
         \end{array},
    \right. 
    \end{align*}
    where $z=(z_1,z_2,z_3)^T \in \mathbb{R}^{3 \times 1}$, and $z_i \sim \text{Normal}(0,1)$ for $i=1,2,3$. Suppose that we get $a^{(2)}$ as the action of the second stage. We  generate the reward as
    \begin{align*} 
  \left\{ 
         \begin{array}{lr}
         r(s,a) =  f_1(s)+ 0.1z' \quad \text{if} \; a^{(2)}=1\\
        r(s,a) =   f_2(s)+ 0.1z'\quad \text{if} \; a^{(2)}=2
         \end{array},
    \right. \qquad \text{where} \; z' \sim \text{Normal}(0,1) \; \text{for} \; i=1,2,3.
    \end{align*}    
    
    To generate the action of the first stage, we introduce the notation,
    \begin{align*}
    g(s^{(1)},a^{(1)}) &= \max\limits_{a^{(2)}}  \mathbb{E}_{r,s}(r(s^{(2)},a^{(2)})) \\
    a^{(1)*} &=  \arg\max\limits_{a^{(1)}} \mathbb{E}_{s} [g(s^{(1)},a^{(1)})]
    \end{align*}    
    where $\mathbb{E}_{r,s}$ is taken over the randomness of the noise of the reward function and of the generation of the state $s^{(2)}$, and $\mathbb{E}_{s}$ is only taken over the randomness of the generation of the state $s^{(2)}$. We generate $a^{(1)}$ with the probability distribution of $P(a^{(1)} = a^{(1)*}) = 1-\epsilon$, where $\epsilon$ is a fixed greedy parameter. 
    
    To generate the action of the second stage, we define
    \begin{align*}
    a^{(2)*} = \arg\max\limits_{a^{(2)}}  \mathbb{E}_r(r(s^{(2)},a^{(2)})),   
    \end{align*}
    where $\mathbb{E}_r$ is taken with respect to the randomness of the reward function. We then generate $a^{(2)}$ with the probability distribution of $P(a^{(2)} = a^{(2)*}) = 1-\epsilon$. 
    
    \item Nonlinear Signal: We define two transformation functions,
    \begin{align*}
    f_1(s) &=  [0.1\exp(4s_1) + 4/(1+\exp(-20(s_2-0.5)))+3s_3+2s_4+s_5]/2.5,  \\ 
    f_2(s) &=  [0.12\exp(4s_1) + 4.8/(1+\exp(-20(s_2-0.5)))+3.6s_3+2.4s_4+1.2s_5]/2.5.
    \end{align*}
    We first randomly generate the coefficient matrix $W_1 \in \mathbb{R}^{2 \times 5}$ with each entry independently drawn from $\text{Normal}(0,1)$. We then define $W_2 = W_1 + 0.05$, where the sum calculation is element-wise. We fix $W_1$ and $W_2$. For each replication, we draw the state at the first stage as $s^{(1)}=(s^{(1)}_1,s^{(1)}_2)^T \in \mathbb{R}^{5 \times 1}$ with $s^{(1)}_i \sim \text{Normal}(0,1)$ for $i=1,2,3,4,5$. Suppose that the action of the first stage is $a^{(1)}$, then we generate the state at the second stage as 
    \begin{align*} 
  \left\{ 
         \begin{array}{lr}
         s^{(2)} =  W_1^{T} s^{(1)} + z \quad \text{if} \; a^{(1)}=1\\
        s^{(2)} =  W_2^{T} s^{(1)} + z \quad \text{if} \; a^{(1)}=2
         \end{array},
    \right. 
    \end{align*}
    where $z=(z_1,z_2,z_3,z_4,z_5)^T \in \mathbb{R}^{3 \times 1}$, and $z_i \sim \text{Normal}(0,1)$ for $i=1,2,3,4,5$.  Suppose that we get $a^{(2)}$ as the action of the second stage. We generate the reward as
    \begin{align*} 
  \left\{ 
         \begin{array}{lr}
         r(s,a) =  f_1(s/10)+ 0.1z' \quad \text{if} \; a^{(2)}=1\\
        r(s,a) =   f_2(s/10)+ 0.1z'\quad \text{if} \; a^{(2)}=2
         \end{array},
    \right. \qquad \text{where} \; z' \sim \text{Normal}(0,1) \; \text{for} \; i=1,2,3,4,5,
    \end{align*}    
    where $s/10$ is calculated in an element-wise fashion. We generate the actions in the same way as for the linear signal.
\end{itemize}

\subsection{Implementation Details and Computing Time} 
\label{appendix_ctime}

For BLBM, we use the RBFSampler function under its default setting in the \textsf{sklearn} package to generate basis with random Fourier features. For BNN, we use a two-layer neural network with 16 hidden units at each layer, and the ReLU activation function. We use SGD for optimization, with the learning rate $10^{-4}$. We set the number of training epochs at 500, the batch size at 100, the number of Monte Carlo samples for gradient descent at 5, and the number of samples from the posterior distribution at 10000. We use \textsf{savio\_htc} cluster for all the computations. For BLBM, it takes about 1.5 seconds to run for one replication on one CPU for the single-stage setting, and about 25 seconds for the two-stage setting. For BNN, it takes about 3 minutes for the single-stage setting, and about 20 minutes for the two-stage setting. We use multiple CPUs for parallelization.

\subsection{Sensitivity Analysis} 
\label{appendix_sensi}

\begin{figure}[t!]
\centering
\includegraphics[width=0.95\textwidth]{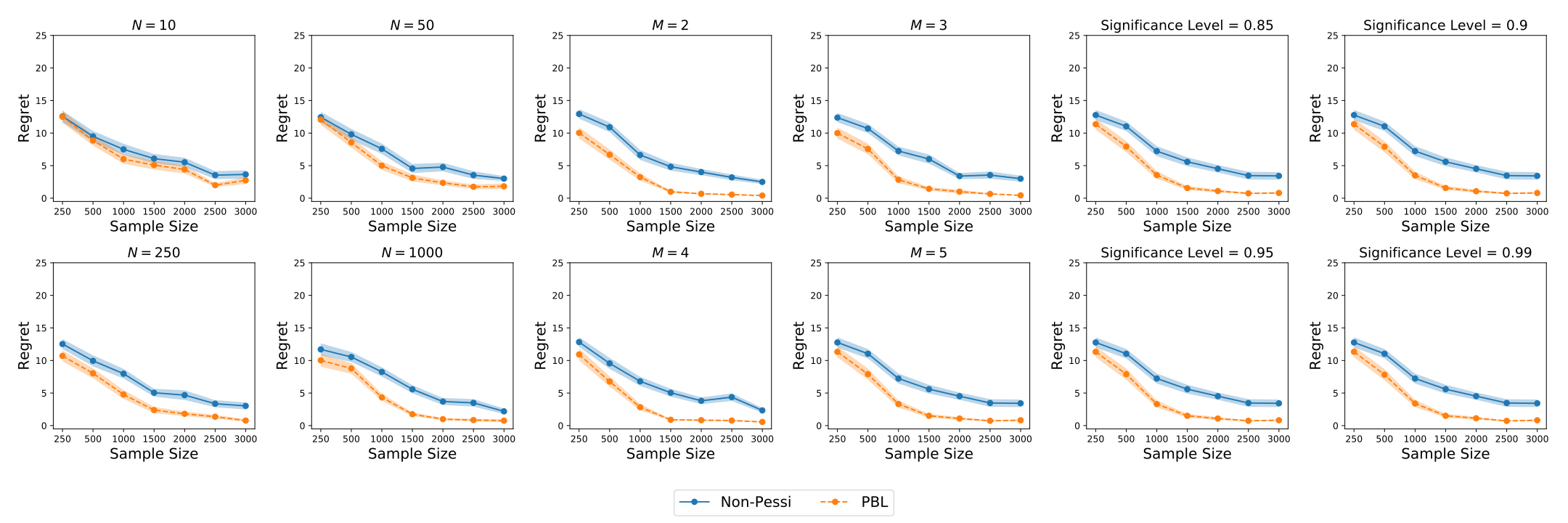}
\caption{Sensitivity analysis for the single-stage contextual bandit simulation with a linear signal.}
\label{fig:sensitivity}
\end{figure}

We conduct a sensitivity analysis by varying a number of parameters in our method, including the number of Monte Carlo samples $N$, the number of ensembles $M$ for the MC gradient computation in variational inference, and the significance level $\alpha$. We adopt the single-stage contextual bandit setting with a linear signal.  Figure \ref{fig:sensitivity} reports the results. It is seen that our method is not overly sensitive to those parameters, as long as they are in a reasonable range. To the contrary, PEVI is sensitive to the choice of the parameter $c$, as shown in Figures \ref{fig:stage1} and \ref{fig:stage2}. 

\subsection{Plot for Two-Stage DTR} 
\label{appendix_plot}

\begin{figure}[t!]
\begin{subfigure}{1\textwidth}
\centering
\includegraphics[width=0.85\textwidth,height=2.4in]{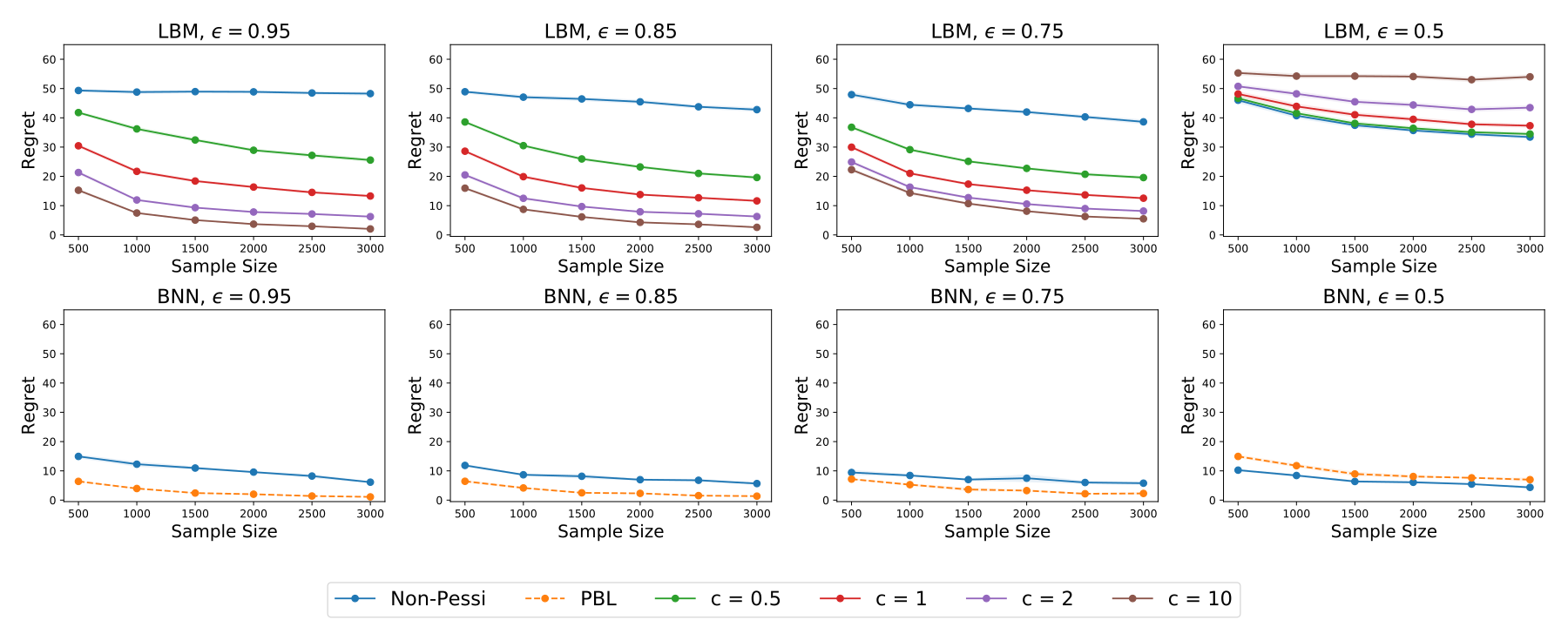}
\caption{Linear Signal}
\label{stage2_linear}
\end{subfigure}
\hspace{1em}
\begin{subfigure}{1\textwidth}
\centering
\includegraphics[width=0.85\textwidth,height=2.4in]{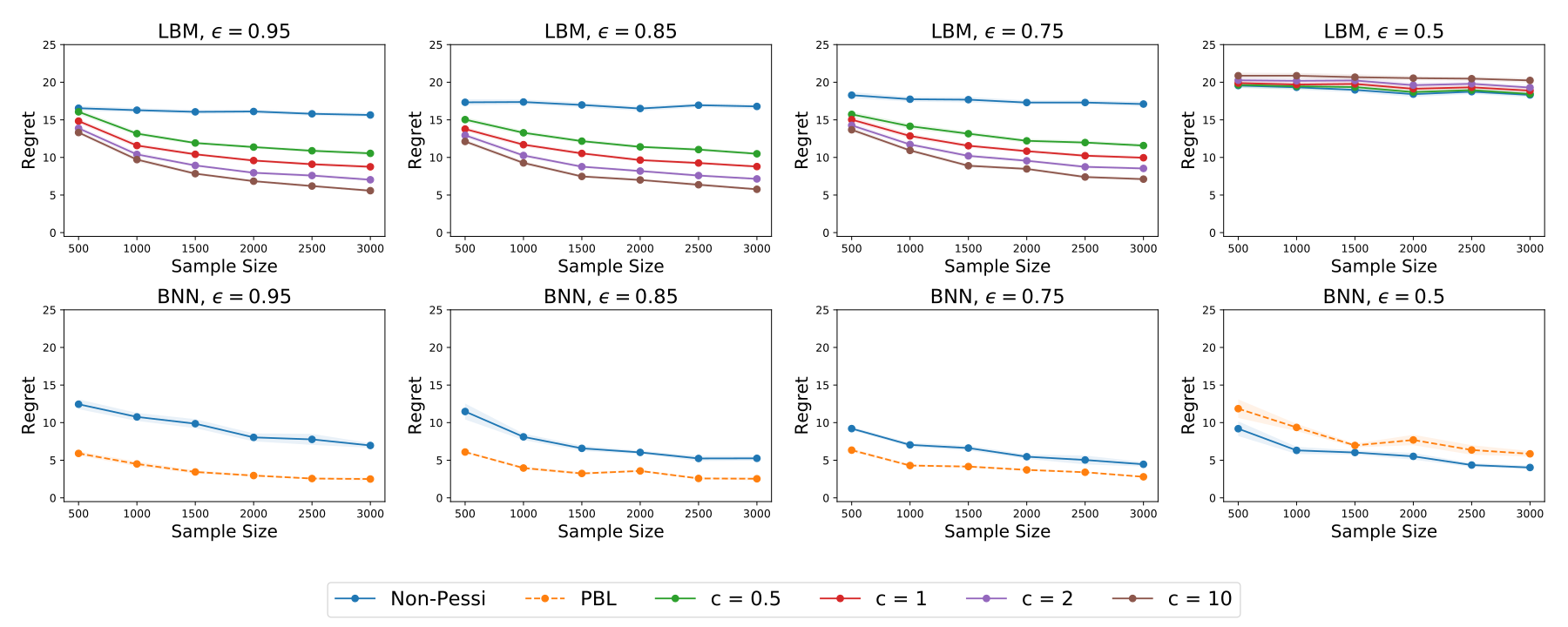}
\caption{Nonlinear Signal}
\label{stage2_nonlinear}
\end{subfigure}
\caption{The two-stage DTR simulation with linear and nonlinear signals. The methods compared include the proposed PBL method, the PEVI method of \cite{jin2021pessimism}, and the standard Q-learning method without pessimism, each of which using BLBM and BNN.}
\label{fig:stage2}
\end{figure}

Figure \ref{fig:stage2} reports the simulation results for the two-stage DTR setting.

\vfill

\end{document}